# Towards French Smart Building Code:

# Compliance Checking Based on Semantic Rules

Nicolas BUS[1], Ana ROXIN[2], Guillaume PICINBONO[1], Muhammad FAHAD[1]

[1] Centre Scientifique et Technique du Bâtiment, Sophia-Antipolis, France
[2] EA7508, CNRS, Arts et Métiers, Univ. Bourgogne Franche-Comté (UBFC), Dijon, France

**Abstract.** Manually checking models for compliance against building regulation is a time-consuming task for architects and construction engineers. There is thus a need for algorithms that process information from construction projects and report non-compliant elements.
Still automated code-compliance checking raises several obstacles. Building regulations are usually published as human readable texts and their content is often ambiguous or incomplete. Also, the vocabulary used for expressing such regulations is very different from the vocabularies used to express Building Information Models (BIM). Furthermore, the high level of details associated to BIM-contained geometries induces complex calculations. Finally, the level of complexity of the IFC standard also hinders the automation of IFC processing tasks.
Model chart, formal rules and pre-processors approach allows translating construction regulations into semantic queries. We further demonstrate the usefulness of this approach through several use cases. We argue our approach is a step forward in bridging the gap between regulation texts and automated checking algorithms. Finally with the recent building ontology BOT recommended by the W3C Linked Building Data Community Group, we identify perspectives for standardizing and extending our approach.

## 1    Introduction

Manually checking models for building code-compliance is time-consuming for architects and construction engineers. According to McGraw-Hill [15], they can spend up to 100 hours on this aspect for complex projects. We, therefore, need algorithms that process construction project information and report non-compliant elements.

Building and construction regulation in France are published as human readable texts. The translation from texts into algorithms is a complex task that requires construction experts to collaborate with computer scientists. Indeed, texts are often ambiguous, incomplete and sometimes contradictory and contain implicit knowledge. Thus it is crucial to correctly parse and prepare such texts, in order to make sure that all implicit knowledge has been made explicit and that all contradictory statements have been correctly interpreted. CSTB (Centre Scientifique et Technique du Bâtiment, French Research Center for Architecture and Construction) started working on this topic way back in 2002 [13]. This resulted in a first mapping between the regulations vocabulary and the IFC vocabulary [8].

Today ISO 16739 is the standard for data sharing in the construction and facility management industries. ISO 16739:2013 or IFC (Industry Foundation Classes) specifies a conceptual data schema and an exchange file format for Building Information Model (BIM) data. The conceptual schema is defined in EXPRESS data specification language [15]. While IFC is the standard for exchanging BIM models, other standards exist for describing building systems or elements. Thus checking a building model against compliance constraints has to take into account the other ways in which building data can be described. Given the level of schema heterogeneity of these standards, semantic Web technologies appear as a solution for delivering semantic interoperability among the diversity of schemas. This has resulted in the publication, by bSI (buildingSmart International) of an OWL ontology version of the IFC standard [8].

In this paper, we thus present an approach based on semantic Web technologies. We argue that it can bring several benefits when applied to building automated regulation compliance checking.

## 2  Related work

### 2.1  Semantic Web Technologies in the context of BIM

In the context of BIM resources modeling has been identified as an interesting approach for achieving information interoperability [16]. Existing IFC-related ontologies were conceived as direct syntax mappings between EXPRESS and OWL languages [20][21]. One of the latest and the most solid implementations of an IFC ontology is ifcOWL approach proposed in (W3C Linked Building Data Community Group, 2014). This version of ifcOWL is also a Candidate Standard [17] for buildingSMART (meaning it is considered as an activity in the process of acquiring international consensus before being submitted to the Standards Committee for a final vote).

### 2.2  Regulation's checking

In the context of the buildingSMART community, the common approach used for verifying IFC files is called Model View Definition (MVD). An MVD "defines a subset of the IFC schema that is needed to satisfy one or many Exchange Require-ments of the AEC industry" [19]. MVDs are serialized in a format called mvdXML [1].

The traditional approach of verifying IFC models using MVDXML is limited as stated in [14] and [18]. Major identified limitations are its limited scope when applying conditions and constraints on several branches of an IFC model, poor geometric analysis of an IFC model, lack of mathematical calculations, and support of only static verification of a model. Given these limitations, several works as [14] and [18] investigated approaches based on logical rules as viable alternatives. Pauwels and al. showed the benefits of using web semantic through a simplified ontology derived from ifcOWL to check construction rules [11]. Fahad et al. studied that building code compliance based on the semantic web rule language has more advantages and cover more functionalities of verification as compared to MVDXML approach [2].

## 3 Method

### 3.1 Addressing schema heterogeneity

Given the diversity of schemas that can be defined for containing building data, we started defining a modeling charter. In our vision, a modeling charter is a document describing how construction project stakeholders agree to model and share data among them. This document describes data exchange processes, modeling best practices, export parameters, nomenclatures and classifications to be used. With all actors of the project sharing the same modeling rules drastically decreases the number of possible schemas. The modeling charter is also used to extend IFC properties with ad-hoc property sets dedicated to specific topics - such as certification level, equipment performances, product impact on environment, etc.

As a future goal, the CSTB aims at providing, in the short term, a national modeling charter core to be used as a common starting point when engineering the modeling plan of a project. We are working in close cooperation with architects, building owner, engineers, audit experts, building quality control agencies and government to reach this goal.

### 3.2 Aligning the Building model with the Regulations ontology

The semantic approach allows describing a building as well as a requirement by using the same atomic fact formalism called triple. In the context of the Semantic Web, a triple represents a statement in the form of "Subject – Predicate – Object", where each part is identified by an URI (Uniform Resource Identifier). Only the object of a triple is an exception to this rule, as it can hold a literal value, and in this case the predicate is called a datatype property. In the case all elements of a triple are identified by URIs, the predicate is called an object property.

The ifcOWL vocabulary [17] includes thousands of classes and properties, very close to building physical components. This vocabulary is not designed to deal with performances, requirements and building functionalities. Thus, if we seek to map this vocabulary to a Regulations ontology (described in section 3.3), we can oversee that the alignment between these two vocabularies cannot be based on a bijective transfer function. Indeed, it is impossible to define a one-to-one mapping between the considered terms. Moreover, some regulatory terms can only be defined with a logical statement aggregating various IFC terms. For instance, in our regulation ontology, a simple regulatory term like "highest storey" is defined by ifcOWL concepts: "IfcBuildingStorey" (ifcOWL term) and "elevation property".

The Regulations ontology is built on top of a simplified version of ifcOWL, extended with complex concepts as pertaining to regulations. Regulation concepts are organized in layers. The ifcOWL vocabulary is the ground layer of the resulting vocabulary pyramid. At the very top of the pyramid, we find the regulation-specific vocabulary. A term of a specific layer is defined by using terms belonging to the layers beneath. This layered approach is flexible as it keeps high-level definitions simple by limiting the number of terms and using cascading rules among layers.

The process we implemented for providing a semantic regulatory view based on an IFC source model comprises several steps. First, we convert the IFC file into a RDF file using the Open-IFC-to-RDF-converter [3] along with custom filtering of classes and relations. The output of this first step is stored in a repository of a triple store. Second, we use a "geometrical pre-processor" that renders and filters geometry aspects. Third, a "semantic preprocessor" filters and enriches the triple store knowledge base with terms from the Regulations ontology.

During this process, some IFC patterns are simplified [11]. For instance, room classification is expressed in ifcOWL with at least ten triples, whereas the Regulations ontology allows narrowing this number down to one triple (see Fig. 1). Other IFC patterns are converted to higher level concepts from the Regulations ontology such as the "highest storey" concept.

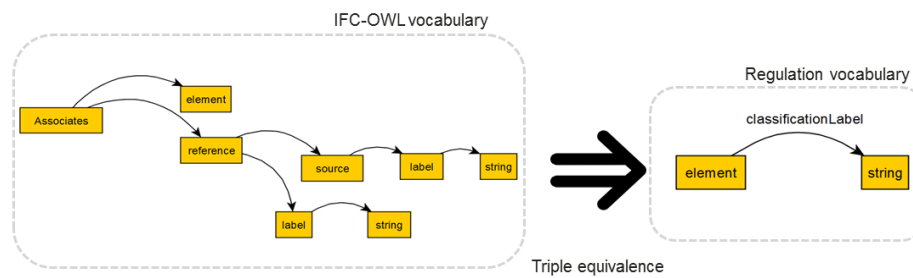

**Fig 1**. Semantic shortcuts for simplifying classification assignment

From a regulatory point of view, the geometric pre-processor only passes significant classes and relations as highlighted in [4] and [12]. Indeed, most of the time a bounding box is sufficient to express regulatory geometrical constrains, instead of relying IFC-based descriptions which contains large sets of tiny facets. Handling such constraints trough a bounding box simplifies their management in the system. Coordinates of each bounding box are expressed by triples.

### 3.3     From regulatory texts to query based constraints

This section provides details about the methodology used to transform the regulatory constraints, expressed in natural language, into processable queries.

First, regulatory texts are prepared and interpreted by working groups of experts. They rely on text editors' auto completion and syntax highlighting tools for re-writing regulatory texts into semi-formal rules. Each semi-formal rule is supposed to report building elements that are non-compliant with regard to a matter or regulation. Depending on its complexity, a regulatory rule expressed in a legal jargon can be divided into several atomic semi-formal rules dealing with complementary aspects. The idea is to keep each semi-formal rule as simple as possible. As a guideline, we suggested that each semi-formal rule begins with an "IF" statement followed by a condition on specific elements and ends with a "THEN NON-COMPLIANT" statement.

The regulatory constraints are pushed from semi-formal to formal stage. This step is performed by computer scientists familiar with BIM and Semantic Web technologies. When semi-formally expressed, rules are easier to adapt into formal rules. Each semi-formal rule is translated into a SPARQL query using the same level of vocabulary coming from the Regulations ontology. Relying on such controlled vocabulary (e.g. the Regulations ontology) when expressing queries allows construction experts to still understand the underlying constraints and rules. Geometrical constraints are described by using geoSPARQL (OGC standard) concepts covering all topological relations that are possible between two geometries as introduced by McGlinn et al [5]. Simplifying the geometry eases the handling of geometry coordinates. This approach should be extended if needed by using functional extensions, as it is the case in [6].

### 3.4     Automating checking

Automating code-compliance checking consists of linking the output of conversion algorithm with the input of geometrical and semantic pre-processors as presented in section 3.1. Each time a new IFC model is submitted to the semantic checker this chain (algorithm + pre-processors) is triggered. The final output is a triple store populated with data with the correct level of details and aligned with the Regulations ontology. The geometrical pre-processor is java-based and returns geometrical triples from IFC. The semantic pre-processor consists of a sequence of semantic forward chaining operations that filters unnecessary information or enhances the knowledge base with high level vocabulary. Indeed, some IFC terms are almost never addressed by regulation texts -such as high-level geometry, element history or sensor states. Thus they can be removed from the knowledge base in order to reduce the amount of data. Ontology alignment and extensions to ifcOWL are declared in the Regulations ontology. SPARQL queries are organized by regulation topics (fire safety, accessibility, airing, acoustics…) so that the architect can focus on a specific aspect. Technically, constraint-queries are packed as SPARQL files in a ZIP archive. The archive also contains metadata for end-users' information purpose. Next steps for further automating algorithm consist in executing, one by one, each formal constraint-query corresponding to each chosen topic and verify the results.

The system outputs a Building Collaboration Format (BCF) file reporting for each constraint-query a list of non-compliant building elements. The BCF file makes it convenient to display such results within a standard IFC viewer. This allows displaying, on top of a graphical representation of a building element, the list of constraints that it breaks.

## 4    Results

We applied this approach on several constraints of the French regulation on both fire safety and accessibility domain. A first version of a French regulation ontology based on previous CSTB research works has been adapted [7]. A group of domain experts had analysed, interpreted and finally converted a dozen of regulation texts into about one hundred semi-formal constraints implementing concepts of the regulatory ontology vocabulary. The regulation ontology was then extended with experts' new concepts suggestions.

The following paragraphs illustrate in detail two specific use cases. According to French building regulation, a WC seat (water closet seat) must have a free space of 0.8 x 1 m at its left side OR at its right side. The "WC" is a high level regulatory concept corresponding to an IfcFlowTerminal with an IfcFlowTerminalType with a predefined type equal to "WCSEAT" [8]. The "FreeSpace" is also a high level regulatory concept. It represents a virtual object (not physical) with 3D dimensions (bounding box). This object shall not intersect any other physical element of the building.

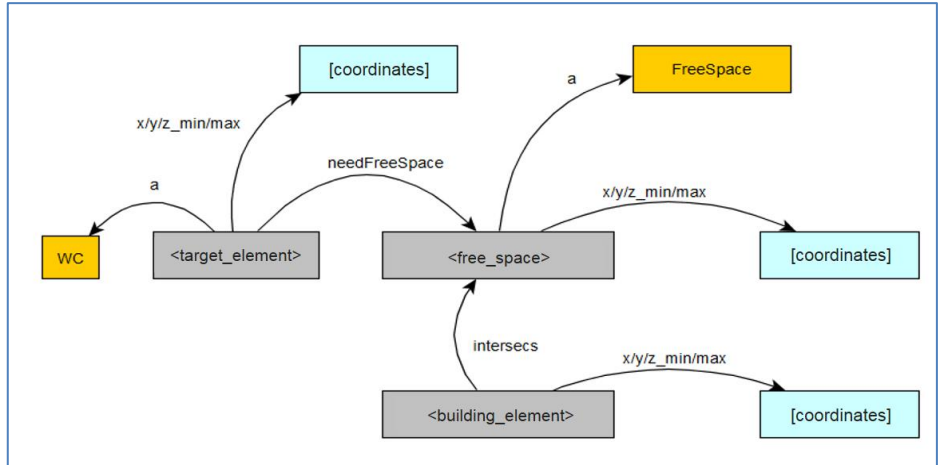

**Fig 2.** "FreeSpace around WC seat" constraint-query as a graph

The checking process of this constraint (see Fig. 2) is detailed by the following sequence:

1. For each "WC" (water closet seat), two "FreeSpace" -and their bounding box- are created on-the-fly. One on the left and one on the right.
2. For each FreeSpace, the query retrieves all building elements that intersects.
3. Finally, the query returns all "WC" with at least one "FreeSpace" intersecting at least one building element.

In this use case, a bounding box representation of a building element is accurate enough to detect clashes with the "WC". Yet, the bounding box must be oriented to be able to locate left and right sides.
In addition to the list of elements that break rules, the explanation query displays the list of elements that intersects with a "FreeSpace". During our test we manage to detect a handrail that was not located on the right side of the "WC".

The fire safety domain regulation put some constraints on the structure element performances. Indeed, from the fire safety perspective, the structure elements of the building must remain bearing load during at least one hour in case of fire. This constraint differs according to the building height. Furthermore, the building height is measured from the ground until the floor of the latest storey. Translated as a SPARQL query, this constraint implies various high-level concepts such as: highest level (storey with highest elevation), storey floor (lower slab of a storey), structure element (building element that bear load), and loadbearing duration in case of fire. The following sequence is triggered by the constraint-query:

1. The building height according to the fire safety standard is calculated on the fly

2. The material performance threshold is determined according to the building height
3. All structure elements are retrieved and their performances are checked
4. Finally, the query returns all structure elements with fire resistance below the threshold.

This architecture has been implemented on the "French national BIM platform" providing a full automated checking process on various fire safety and accessibility topics.

## 5  Conclusion

As compared to MVDXML, we gained several advantages from our Semantic checker approach. At a first look, it needs transformation of an IFC model to the RDF representation which is an overhead and time taking process. But, once the semantic repository is prepared, it provides ease to unify, explore, analyze, and extend triples for the verification of IFC models. We conclude that semantic technologies provide more rich mechanisms and answer vast types of queries for the verification of IFC models. It provides powerful features based on SPARQL libraries and serves best for the automated code compliance and verification of IFC models.

While translating regulatory text to formal constraint-queries, we introduced a semi-formal stage that is readable by both construction experts and data scientists. This study also points out that aligning the model to high level vocabulary makes it easier to write and maintain constraint-queries.

Semantic techniques make it possible to translate IFC terms to high level through two different way, i.e., forward chaining or backward chaining. Backward chaining consists of ontology statements that align IFC concepts with regulatory concepts, whereas forward chaining consists in creating supplementary triples. While, backward chaining using ontology statements is theoretically the best approach, we found that forward chaining is a good alternative at an optimization stage to save memory and CPU resources. From the machine point of view, backward chaining is processed each time a semantic query is submitted whereas forward chaining is executed each time the data changes.

We found that constraints included in regulations texts mostly consist in retrieving elements and checking their properties. Some constraints concerning geometry do not require a high level of details. Most frequent topological relations expressed in constraints deal with the intersection and adjacency. We also found that some information required when considering regulatory aspects are missing in the IFC specification. Some properties specific to national regulation need to be added to the IFC schema. Fire escape plan is also out of the IFC scope but is required when checking doors opening direction.

Obviously, this approach does not address all constraints defined in the regulations. For instance, when considering a complex corridor with multiples aisles -and their different related widths - the bounding box approximation will output erroneous results while computing intersection between spaces. Nevertheless, we argue that this approach can help in addressing the issues mentioned above. Furthermore, we envision extending it by considering the latest works done in the Linked Building Data Community group such as the Building Topology Ontology (BOT [9]), the ontology defining the core concepts of a building and OntoBREP [10] - ontology for CAD Data and Geometric Constraints.